\documentclass[conference]{IEEEtran}
\IEEEoverridecommandlockouts

\usepackage{cite}
\usepackage{amsmath,amssymb,amsfonts}
\usepackage{algorithmic}
\usepackage{graphicx}
\usepackage{textcomp}
\usepackage{xcolor}
\usepackage{booktabs}
\usepackage{array}
\usepackage{float}
\usepackage[export]{adjustbox}

\def\BibTeX{{\rm B\kern-.05em{\sc i\kern-.025em b}\kern-.08em
    T\kern-.1667em\lower.7ex\hbox{E}\kern-.125emX}}
    
\begin{document}

\title{Enhancing Standard and Dialectal Frisian ASR: Multilingual Fine-tuning and Language Identification for Improved Low-resource Performance\\
}


\author{
\IEEEauthorblockN{Reihaneh Amooie}
\IEEEauthorblockA{\textit{Center for Language and Cognition} \\
\textit{University of Groningen}\\
Groningen, The Netherlands \\
r.amooie@rug.nl}
\and
\IEEEauthorblockN{Wietse de Vries}
\IEEEauthorblockA{\textit{Center for Language and Cognition} \\
\textit{University of Groningen}\\
Groningen, The Netherlands\\
wietse.de.vries@rug.nl}
\and
\IEEEauthorblockN{Yun Hao}
\IEEEauthorblockA{\textit{Center for Language and Cognition} \\
\textit{University of Groningen}\\
Groningen, The Netherlands \\
yun.hao@rug.nl} 
\and 
\IEEEauthorblockN{Jelske Dijkstra}
\IEEEauthorblockA{\textit{Mercator European Research Centre} \\
\textit{Fryske Akademy}\\
Friesland, The Netherlands \\
jdijkstra@fryske-akademy.nl\\
}  
\and
\IEEEauthorblockN{Matt Coler}
\IEEEauthorblockA{\textit{Speech Technology Lab} \\
\textit{University of Groningen}\\
Friesland, The Netherlands \\
m.coler@rug.nl}
\and
\IEEEauthorblockN{Martijn Wieling}
\IEEEauthorblockA{\textit{Center for Language and Cognition} \\
\textit{University of Groningen}\\
Groningen, The Netherlands \\
m.b.wieling@rug.nl}
}

\maketitle
\begin{abstract}
Automatic Speech Recognition (ASR) performance for low-resource languages is still far behind that of higher-resource languages such as English, due to a lack of sufficient labeled data.  State-of-the-art methods deploy self-supervised transfer learning where a model pre-trained on large amounts of data is fine-tuned using little labeled data in a target low-resource language. In this paper, we present and examine a method for fine-tuning an SSL-based model in order to improve the performance for Frisian and its regional dialects (Clay Frisian, Wood Frisian, and South Frisian). We show that Frisian ASR performance can be improved by using multilingual (Frisian, Dutch, English and German) fine-tuning data and an auxiliary language identification task. In addition, our findings show that performance on dialectal speech suffers substantially, and, importantly, that this effect is moderated by the elicitation approach used to collect the dialectal data. Our findings also particularly suggest that relying solely on standard language data for ASR evaluation may underestimate real-world performance, particularly in languages with substantial dialectal variation.
\end{abstract}


\section{Introduction}
Automatic Speech Recognition (ASR) systems are an integral component in human-machine interaction systems. 
Despite  advancements driven by the availability of extensive data and computational resources, achieving near human-like performance remains confined mainly to high-resource languages and standard dialects, leaving minority languages underrepresented. 
Our study  approaches this challenge within the context of Frisian, a minority language from the province of Friesland, The Netherlands, and its regional dialects: Klaaifrysk (Clay Frisian), Wâldfrysk (Wood Frisian), and Súdwesthoeksk (South West Frisian, hereafter referred to as South Frisian). 
Frisian and its varieties present a unique and compelling case study for low-resource ASR research due to several factors. As a minority language closely related to Dutch and German, Frisian exists in a complex multilingual environment where code-switching and interference from dominant languages are common. Moreover, as mentioned, Frisian encompasses several distinct dialects, each with its own phonological and lexical variations. This dialectal diversity within a single low-resource language creates additional challenges for ASR. By addressing these challenges, our work has broader implications for other low-resource languages with similar challenges. 
Furthermore, this research contributes to the preservation and technological support of minority languages, an important consideration in the field of speech technology.

In light of the potential for multilingual training to enhance ASR performance across diverse linguistic landscapes~\cite{chen2023improving}, our research, which is inspired by the works of Chen and colleagues  \cite{chen2023improving}, and Liu and colleagues \cite{liu2023hierarchical}, explores the application of multilingual self-supervised models. In that context, our study aims to determine the effectiveness of leveraging multilingual fine-tuning data alongside a targeted strategy (language identification) during fine-tuning, to adapt the advanced SSL models to the specific needs of Frisian and its dialects. 




\section{Related Work}
\label{sec:format}

Various approaches have been proposed to improve the performance of low-resource ASR systems. 
One important line of research focuses on the potential of transferring cross-lingual knowledge from high-resource to low-resource languages~\cite{yadav2022survey}. Recent work has found that multilingual models can be beneficial in improving ASR performance for low-resource languages, as they can learn universal features that are transferable across  languages. Specifically, Liu and colleagues~\cite{liu2023hierarchical} proposed a method for explicit transfer of cross-lingual knowledge at the decoding stage. Their proposed model (which performs multilingual hierarchical Softmax decoding using a Huffman tree) can capture cross-linguistic similarity among units such as characters and phones. 
However, despite being relatively  effective, these types of cross-lingual transfer approaches introduce further challenges as a result of variability in phonology, grammar, and orthography across different languages~\cite{chen2023improving}.
A promising avenue for improving and optimizing multilingual end-to-end ASR systems is making the models aware of the identity of the languages and dialects that they should be transcribing~\cite{kashiwagi2024rapid}. The most basic method for this is implementing a language identification (LID) module as a pre-processing step before ASR processing~\cite{kwon2023mole}. This, however, results in higher latency~\cite{waters2019leveraging}.  
A possible technique to address this latency issue is jointly performing the two tasks of language identification and speech recognition via multi-task learning \cite{zhang2022streaming}. 
In a related approach, Kwon and Chung \cite{kwon2023mole} implemented a system containing multiple language experts (MoLEs; adapted from Mixture-of-Experts  \cite{you2022speechmoe2} together with a Switch Transformer \cite{fedus2022switch}). This system is then deployed to assign the input sequence (a sentence) to a certain language-specific expert according to the language type, and the language-agnostic experts process it regardless of language. 
An alternative strategy involves the technique proposed by Chen and colleagues  \cite{chen2023improving}, where an auxiliary CTC~\cite{graves2012connectionist} objective enables the early encoder layers to focus on language identification, so that, 
the later layers can be conditioned on the predicted language. 
Inspired by the above-mentioned approaches, in this work, we propose combining multilingual fine-tuning data and language identification for improving ASR for Frisian and its regional varieties.

\section{Method}
\label{sec:pagestyle}

\subsection{Data}
\subsubsection{Fine-tuning data}
\label{sssec:subsubhead}
As a first step in our approach, we fine-tune our pre-trained base model with monolingual Common Voice (CV) 17.0 \cite{commonvoice:2020} Frisian data. Specifically, we use the official training split data for this purpose, which consists of about 5.5 hours of validated read speech from 195 different speakers. The source material was presented in Standard Frisian. This model serves as our baseline.  

In a second step, we create multilingual fine-tuning data by adding Common Voice 17.0 data from three other  West Germanic languages, namely Dutch, German and English. We create separate fine-tuning datasets by incrementally adding more languages to the monolingual Frisian dataset. The order in which the languages were added to the fine-tuning data was based on their similarity to Frisian. The similarity was calculated based on 
lexical-phonetic distances of Frisian, Dutch, German and English on the basis of the Automated Similarity Judgement Program (ASJP) data \cite{wichmann2010evaluating}, and was adapted from \cite{de2021adapting}. Based on this similarity measure, we first added Dutch (being the most similar to Frisian), then German, and finally English.\footnote{In line with historical developments of Frisian, we also experimented with adding English before German, but this resulted in lower performance compared to only including Frisian and Dutch data.} The audio data for all languages was sampled at 16 kHz. In order to make the multilingual fine-tuning datasets more balanced and prevent bias, the size of the original datasets for Dutch, German and English were each reduced by random sampling to that of the  Frisian dataset (3,921 sentences). Table~\ref{tab:finetuning_data_info} provides an overview of the resulting fine-tuning datasets.



\begin{table*}[t]
\caption{Datasets used for fine-tuning (subsets from CV 17.0~\cite{commonvoice:2020}).}
\begin{center}
\begin{tabular}{|l|l|c|c|}
\hline
\textbf{Languages} & \textbf{ISO Codes} & \textbf{Duration} & \textbf{Sentences} \\
\hline
Frisian & fy & 5.5h & 3,921 \\
\hline
Frisian, Dutch & fy-nl & 11.0h & 7,842 \\
\hline
Frisian, Dutch, German & fy-nl-de & 16.0h & 11,763 \\
\hline
Frisian, Dutch, German, English & fy-nl-de-en & 22.0h & 15,684 \\
\hline
\multicolumn{4}{l}{}
\end{tabular}
\label{tab:finetuning_data_info}
\end{center}
\end{table*}

\subsubsection{Development data}
We used the development split from Common Voice 17.0 as our development dataset. This dataset  contains 3,170 sentences uttered by 237 speakers based on Standard Frisian text. The audio data was sampled at 16 KHz.  
\subsubsection{Evaluation data}
\label{sssec:subsubhead}
\begin{figure}[h]
   \centering
   \includegraphics[width=0.5\textwidth]{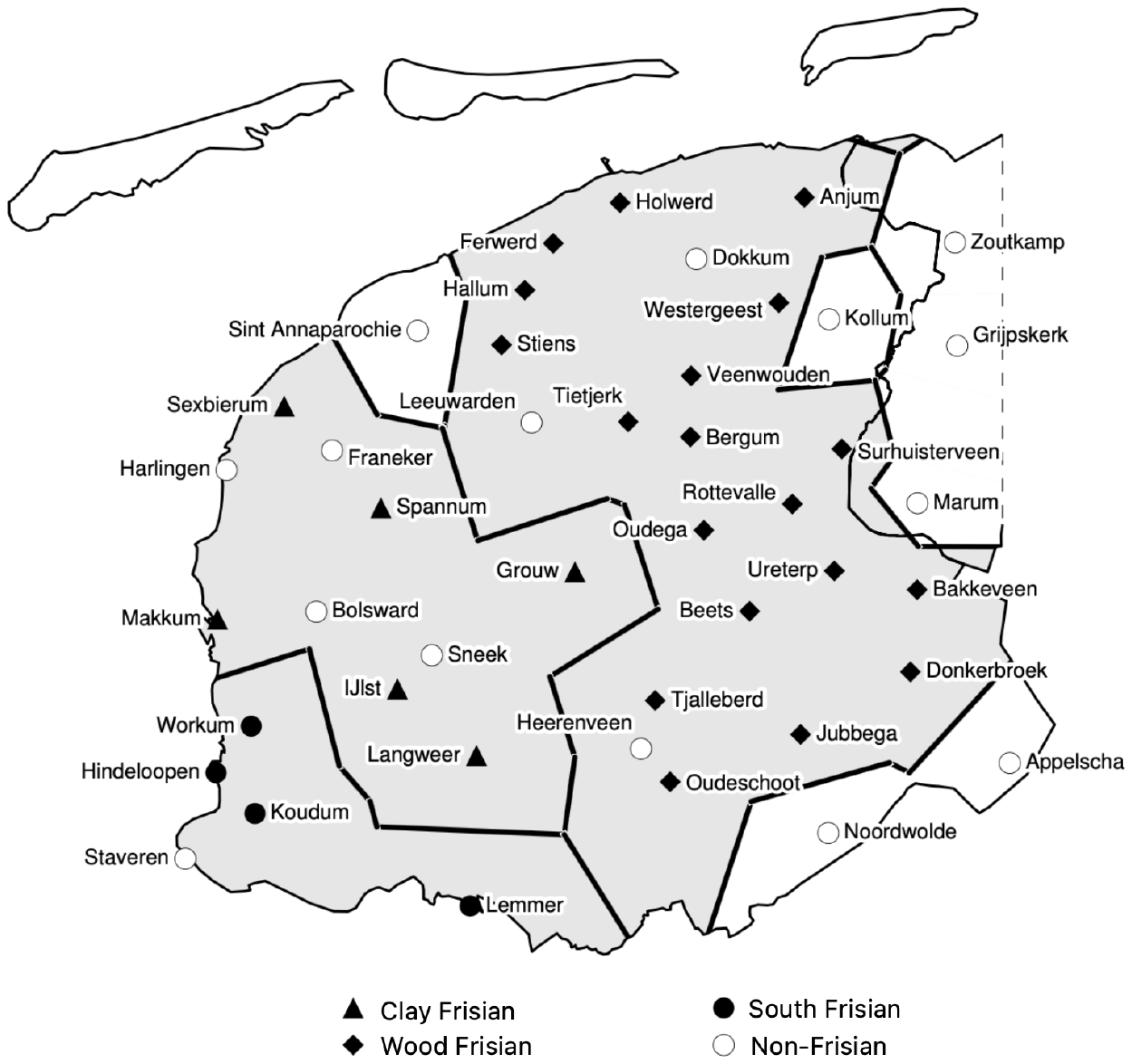}
   \caption{Distribution of Frisian dialects spoken in the province of Friesland, the Netherlands. Figure adapted from~\cite{heeringa2005dialect}.} 
   \label{fig:dialects}
\end{figure}

In order to evaluate the resulting fine-tuned models, we use two different datasets. First, we use the official Common Voice 17.0 Frisian test split which includes 3,171 sentences of read speech based on Standard Frisian.  This enables us to examine the performance of the models on Standard Frisian speech. Next, we investigate the performance on dialectal speech. In order to collect dialectal Frisian speech, we used the SPRAAKLAB \cite{wieling2023spraaklab} mobile laboratory to collect dialectal speech data from three regions in the province of Friesland, where different dialects are spoken (see Figure~\ref{fig:dialects}). The collected SPRAAKLAB data can be divided into two subsets. 
For the first subset, the speakers were shown a group of sentences in Dutch (the national language of the Netherlands which is taught in schools). This text was translated from an original Frisian story book.\footnote{https://websjop.afuk.frl/frl/winkel/nee-heb-je-ja-kunje-krijgen-het-verhaal-van-de-lytse-yerke/} The speakers were asked to translate each sentence in their head, after which they uttered the sentence in their own regional variety (Clay, Wood, or South Frisian). The utterances were then manually transcribed and used as labels in our evaluation dataset. In the second subset, the participants were shown the original Standard Frisian sentences from the same story book and they uttered them in their own regional variety. The reason for this division is that translating from Dutch text to spoken dialectal Frisian allows the speech to be potentially less constrained by Standard Frisian norms, which may result in more natural dialectal variation. The group of participants comprised of sixteen men, all above the age of 60. We balanced the groups, such that both modalities had an equal number of speakers per region. Consequently, in both groups there were three Wood Frisian speakers, two Clay Frisian speakers and three South Frisian speakers. Both the subset collected based on Dutch text and the subset collected based on Standard Frisian text contains 824 sentences (103 sentences per speaker). All audio recordings were sampled at 16KHz. 

\subsection{Experimental Setup}

\subsubsection{Model Architecture}

As our pre-trained base model, we used the XLS-R 1B architecture \cite{babu2021xls}, which contains 1 billion parameters and was trained on 436,000 hours of speech data from 128 languages. This model utilizes a Transformer \cite{vaswani2017attention} architecture with 48 encoder layers, each containing 1,024 hidden units and 16 attention heads. 

\subsubsection{Fine-Tuning Process}
 For fine-tuning, we froze the feature extractor and only updated the encoder layers. We used the following hyperparameters:
{\begin{itemize}
    \item Learning rate: 0.00008
    \item Batch size: 8 with 16 gradient accumulation steps
    \item Maximum number of training epochs: 50
    \item Loss function: CTC loss
    \item Vocabulary size: 87
    
\end{itemize}}
All experiments were conducted using 16-bit floating point precision on a single NVIDIA A100 GPU. We used a staged approach for fine-tuning:
\begin{itemize}
    \item \textbf{Monolingual baseline:} We initially fine-tuned the model using only the Common Voice 17.0 Frisian dataset to establish the baseline performance.
    \item \textbf{Incremental multilingual integration:} We progressively introduced additional languages (Dutch, German, and English) to the fine-tuning data. The order of language addition was based on their linguistic proximity to Frisian, determined by the LDND distances. 
    In other words, after starting with the model based only on Frisian fine-tuning data, a second model was fitted with Frisian and Dutch fine-tuning data. The third model was fitted with Frisian, Dutch and German fine-tuning data, and the final model was fitted using  fine-tuning data from all four languages. To prevent bias towards higher-resource languages, we down-sampled the data from Dutch, German, and English to match the number of sentences available in the Frisian data.
    
    \item \textbf{LID token integration:} To make the model aware of the language it should transcribe, we included language identification in the multilingual models. 
    We implemented this 
    by adding an LID token at the beginning of each target transcription. This approach allows the model to learn to recognize the language and adjust its processing based on the language context~\cite{waters2019leveraging}.   
\end{itemize}

\begin{table*}[htbp]  
\caption{The impact of using multilingual fine-tuning data on standard test data. WER scores are only based on Frisian (CV 17.0) test data.}
\begin{center}  
\begin{tabular}{|l|c|c|}
\hline
 \textbf{Fine-tuning language(s)} & \textbf{Dev WER} & \textbf{Test WER} \\
\hline
 fy        & 13.4   & 14.2  \\
\hline
 fy-nl     & 13.2   & 13.6  \\
\hline
 fy-nl-de  &  \textbf{12.6} & \textbf{13.1} \\
\hline
fy-nl-de-en & 12.9  & 13.4   \\
\hline
\end{tabular}
\label{tab:data-info-1}
\end{center}  
\end{table*}

\begin{table*}[htbp]  
\caption{The impact of including language identification tokens during fine-tuning on the WER of models fine-tuned on multilingual data. WER scores are only based on CV 17.0 Frisian test data. Similarly, we only report LID recall for Frisian since we only test on Frisian.}
\begin{center}  
\begin{tabular}{|l|c|c|c|}
\hline
\textbf{Fine-tuning languages} & \textbf{LID recall} & \textbf{Dev WER} & \textbf{Test WER} \\
\hline
fy-nl   & - & 13.2 & 13.6 \\       
\hline
 fy-nl + LID & 99.7 & 12.8 & 13.1 \\
\hline
 fy-nl-de & - & 12.6 & 13.1 \\
\hline
 fy-nl-de + LID & 99.3 & \textbf{12.6} & \textbf{13.1} \\
\hline
fy-nl-de-en & - & 12.9 & 13.4 \\
\hline
fy-nl-de-en + LID & 99.4 & 13.0 & 13.2 \\
\hline
\end{tabular}
\label{tab:lid_impact}
\end{center}  
\end{table*}

\begin{table*}[htbp]  
\caption{The impact of using multilingual fine-tuning data and language identification on dialectal Frisian test data. Two WERs are shown, the subscripts denote the language in which the original sentences were presented (SF: Standard Frisian; D: Dutch).}
\begin{center}  
\begin{tabular}{|l|c|c|c|c|}
\hline
\textbf{Fine-tuning languages} & \textbf{LID$_\textrm{SF}$ recall} & \textbf{LID$_\textrm{D}$ recall} & \textbf{WER$_\textrm{SF}$} & \textbf{WER$_\textrm{D}$} \\
\hline
 fy & - & - & 25.1 & 26.0 \\
\hline
 fy-nl & - & - & 27.0 & 26.2 \\
\hline
 fy-nl + LID & 97.5 & 99.3 & 25.1 & 25.2 \\
\hline
 fy-nl-de & - & - & 25.6 & 26.3 \\
\hline
 fy-nl-de + LID & 95.5 & 98.5 & \textbf{24.1} & 25.4 \\
\hline
fy-nl-de-en & - & - & 26.2 & 25.9 \\
\hline
 fy-nl-de-en + LID & 87.3 & 91.6 & 24.9 & \textbf{25.1} \\
\hline
\end{tabular}
\label{tab:data-info-2}
\end{center}  
\end{table*}

\section{Results and Discussion}
\label{sec:typestyle}
In this section, we report and discuss the results of our experiments.

\textbf{Monolingual vs.~ multilingual fine-tuning data:}
We report word error rate (WER (\%)) for three different models and compare their performance with our baseline. In Table~\ref{tab:data-info-1}, the WER is reported for Common Voice 17.0 development and test sets. Compared with the model fine-tuned on monolingual Frisian data, adding both Dutch and German languages to the fine-tuning data appears to improve the performance by  1.1\% (a relative reduction in error of about 7.5\%). However, adding English data appears to be detrimental, as it instead results in diminished performance (13.4\%). This may be due to the larger distance between English and Frisian. 

\textbf{Language identification:}
We observe only minor improvements in WER of the LID addition in the CV 17.0 experiments with one, two, and three additional languages (Frisian-Dutch,  Frisian-Dutch-German, and Frisian-Dutch-German-English), as detailed in Table~\ref{tab:lid_impact}. The best performance for Frisian was achieved with the multilingual model including Frisian, Dutch and German data, regardless of the use of LID tokens. However, a more pronounced beneficial effect of LID tokens was observed when evaluating using the dialectal SPRAAKLAB data. Here, the inclusion of LID tokens led to a reduction in WER of approximately 1\% (see Table~\ref{tab:data-info-2}). The greater benefit of LID in this case may be due to the relatively higher WER for this dataset. 
\newline
\textbf{Standard vs. dialectal speech:}
To compare the performance of our models on standard and dialectal speech, we evaluate the models using the SPRAAKLAB  data (see Table~\ref{tab:data-info-2}). We evaluate all models on the datasets with Frisian and Dutch stimuli and compare their performance with our baseline. As expected, all of the models showed poorer performance on dialectal data, which suggests that the Common Voice data is relatively close to Standard Frisian. The higher performance on the Common Voice test set may also be attributed to the fact that CV test dataset is more similar to its training subset than the SPRAAKLAB data, regardless of the dialect of its speakers. Moreover, we observed that the beneficial effect of adding Dutch and German to the fine-tuning data is mostly consistent with the previous experiments on standard speech, but only in combination with LID and not without. Additionally, the models perform slightly better on the data with Standard Frisian stimuli than on the data with Dutch stimuli. This suggests that using Dutch texts (which need to be translated) likely allows for more regional variability in Frisian than using standard Frisian texts.



\section{Conclusion}
\label{sec:print} 
In this paper, we studied the impact of using multilingual fine-tuning data and language identification  
in an ASR model for Frisian. 
We implemented 
the models by incrementally adding more languages from three West Germanic languages 
and incorporating LID tokens at the input. 
We observed that adding Dutch and German helps Frisian ASR performance. Similarly, including language identification information appears to have an additional marginal positive effect. 
We also 
showed that the models performed better on standard speech than dialectal speech. Moreover, 
the experiments indicated that speech is more ``dialectal'' when 
elicited by \textit{not} presenting texts in the standard language. This may have implications for developing ASR systems for minority languages, as performance estimations on the basis of test sets which use stimuli presented in the standard language may underestimate the real-life performance when there is much dialect variation. From a broader perspective, our findings highlight the importance of incorporating shared features among languages in multilingual environments and the potential of context-aware modeling for improving generalization and robustness in ASR systems.

\section*{Acknowledgment}
This work was supported by the Provinsje Frysl\^an. We are also grateful to the ILSE project of the Center for Information Technology, University of Groningen, for
providing access to the computing resources utilized in this research.

\newpage
\bibliographystyle{IEEEbib}
\bibliography{strings,refs}

\begin{thebibliography}{10}

\bibitem{chen2023improving}
William Chen, Brian Yan, Jiatong Shi, Yifan Peng, Soumi Maiti, and Shinji Watanabe,
\newblock ``Improving massively multilingual asr with auxiliary ctc objectives,''
\newblock in {\em ICASSP 2023-2023 IEEE International Conference on Acoustics, Speech and Signal Processing (ICASSP)}. IEEE, 2023, pp. 1--5.

\bibitem{liu2023hierarchical}
Qianying Liu, Zhuo Gong, Zhengdong Yang, Yuhang Yang, Sheng Li, Chenchen Ding, Nobuaki Minematsu, Hao Huang, Fei Cheng, Chenhui Chu, et~al.,
\newblock ``Hierarchical softmax for end-to-end low-resource multilingual speech recognition,''
\newblock in {\em ICASSP 2023-2023 IEEE International Conference on Acoustics, Speech and Signal Processing (ICASSP)}. IEEE, 2023, pp. 1--5.

\bibitem{yadav2022survey}
Hemant Yadav and Sunayana Sitaram,
\newblock ``A survey of multilingual models for automatic speech recognition,''
\newblock {\em arXiv preprint arXiv:2202.12576}, 2022.

\bibitem{kashiwagi2024rapid}
Yosuke Kashiwagi, Hayato Futami, Emiru Tsunoo, Siddhant Arora, and Shinji Watanabe,
\newblock ``Rapid language adaptation for multilingual e2e speech recognition using encoder prompting,''
\newblock {\em arXiv preprint arXiv:2406.12611}, 2024.

\bibitem{kwon2023mole}
Yoohwan Kwon and Soo-Whan Chung,
\newblock ``Mole: Mixture of language experts for multi-lingual automatic speech recognition,''
\newblock in {\em ICASSP 2023-2023 IEEE International Conference on Acoustics, Speech and Signal Processing (ICASSP)}. IEEE, 2023, pp. 1--5.

\bibitem{waters2019leveraging}
Austin Waters, Neeraj Gaur, Parisa Haghani, Pedro Moreno, and Zhongdi Qu,
\newblock ``Leveraging language id in multilingual end-to-end speech recognition,''
\newblock in {\em 2019 IEEE Automatic Speech Recognition and Understanding Workshop (ASRU)}. IEEE, 2019, pp. 928--935.

\bibitem{zhang2022streaming}
Chao Zhang, Bo~Li, Tara Sainath, Trevor Strohman, Sepand Mavandadi, Shuo-Yiin Chang, and Parisa Haghani,
\newblock ``Streaming end-to-end multilingual speech recognition with joint language identification,''
\newblock {\em arXiv preprint arXiv:2209.06058}, 2022.

\bibitem{you2022speechmoe2}
Zhao You, Shulin Feng, Dan Su, and Dong Yu,
\newblock ``Speechmoe2: Mixture-of-experts model with improved routing,''
\newblock in {\em ICASSP 2022-2022 IEEE International Conference on Acoustics, Speech and Signal Processing (ICASSP)}. IEEE, 2022, pp. 7217--7221.

\bibitem{fedus2022switch}
William Fedus, Barret Zoph, and Noam Shazeer,
\newblock ``Switch transformers: Scaling to trillion parameter models with simple and efficient sparsity,''
\newblock {\em Journal of Machine Learning Research}, vol. 23, no. 120, pp. 1--39, 2022.

\bibitem{graves2012connectionist}
Alex Graves and Alex Graves,
\newblock ``Connectionist temporal classification,''
\newblock {\em Supervised sequence labelling with recurrent neural networks}, pp. 61--93, 2012.

\bibitem{commonvoice:2020}
R.~Ardila, M.~Branson, K.~Davis, M.~Henretty, M.~Kohler, J.~Meyer, R.~Morais, L.~Saunders, F.~M. Tyers, and G.~Weber,
\newblock ``Common voice: A massively-multilingual speech corpus,''
\newblock in {\em Proceedings of the 12th Conference on Language Resources and Evaluation (LREC 2020)}, 2020, pp. 4211--4215.

\bibitem{wichmann2010evaluating}
S{\o}ren Wichmann, Eric~W Holman, Dik Bakker, and Cecil~H Brown,
\newblock ``Evaluating linguistic distance measures,''
\newblock {\em Physica A: Statistical Mechanics and its Applications}, vol. 389, no. 17, pp. 3632--3639, 2010.

\bibitem{de2021adapting}
Wietse De~Vries, Martijn Bartelds, Malvina Nissim, and Martijn Wieling,
\newblock ``Adapting monolingual models: Data can be scarce when language similarity is high,''
\newblock {\em arXiv preprint arXiv:2105.02855}, 2021.

\bibitem{wieling2023spraaklab}
Martijn Wieling, Teja Rebernik, and Jidde Jacobi,
\newblock ``Spraaklab: a mobile laboratory for collecting speech production data,''
\newblock in {\em Proceedings of the 20th International Congress of Phonetic Sciences}. Guarant International, 2023, pp. 2060--2064.

\bibitem{heeringa2005dialect}
Wilbert Heeringa,
\newblock ``Dialect variation in and around frisia; classification and relationships,''
\newblock {\em Us Wurk}, vol. 54, no. 3-4, pp. 125--167, 2005.

\bibitem{babu2021xls}
Arun Babu, Changhan Wang, Andros Tjandra, Kushal Lakhotia, Qiantong Xu, Naman Goyal, Kritika Singh, Patrick Von~Platen, Yatharth Saraf, Juan Pino, et~al.,
\newblock ``Xls-r: Self-supervised cross-lingual speech representation learning at scale,''
\newblock {\em arXiv preprint arXiv:2111.09296}, 2021.

\bibitem{vaswani2017attention}
A~Vaswani,
\newblock ``Attention is all you need,''
\newblock {\em Advances in Neural Information Processing Systems}, 2017.

\end{thebibliography}

\end{document}